\documentclass{article} 
\usepackage{arxiv}

\usepackage{url,hyperref,lineno,microtype,subcaption}
\usepackage{graphicx}
\usepackage[onehalfspacing]{setspace}
\usepackage{amsmath}
\usepackage{amsfonts} 
\usepackage{amsthm}
\usepackage{tabularx}

\usepackage[backend=biber, style=numeric]{biblatex}
\title{Initialization of Latent Space Coordinates via \\Random Linear Projections for Learning Robotic Sensory-Motor Sequences}

\author{Vsevolod Nikulin  \\
Cognitive Neurorobotics Research Unit\\
Okinawa Institute of Science and Technology\\
Okinawa, Japan \\
\and
Jun Tani\thanks{Correspondence: \texttt{jun.tani@oist.jp}} \\
Cognitive Neurorobotics Research Unit\\
Okinawa Institute of Science and Technology\\
Okinawa, Japan\\
}

\newtheorem{thm}{Theorem}[]
\DeclareMathOperator{\EX}{\mathbb{E}}

\begin{document}

\maketitle

\begin{abstract}

Robot kinematics data, despite being a high dimensional process, is highly correlated, especially when 
considering motions grouped in certain primitives. These almost linear correlations within primitives allow us to
interpret the motions as points drawn close to a union of low-dimensional linear subspaces in the space of all motions.
Motivated by results of embedding theory, in particular, generalizations of Whitney embedding theorem, we show that
random linear projection of motor sequences into low dimensional space loses very little information about structure of
kinematics data. Projected points are very good initial guess for values of latent variables in generative model for robot 
sensory-motor behaviour primitives. 
We conducted series of experiments where we trained a recurrent neural network to generate sensory-motor sequences for robotic manipulator
with 9 degrees of freedom. Experimental results demonstrate substantial improvement in generalisation abilities for unobserved samples in the case of initialization of latent variables with random linear projection of motor data over initialization with zero or random values. 
Moreover, latent space is well-structured wherein samples belonging to different primitives are well separated from the onset of training process.
\end{abstract}

\section{Introduction}

Generative models allow to represent high-dimensional behaviour patterns (sequences of action-perception pairs) 
in much lower-dimensional latent space. However, these representations are far from unique. It is convenient for analysis and theoretical
arguments about generalization capabilities if encoded points exhibit some regularities. 
The current paper focuses on the issue of efficient encoding of motion primitives. 
It is known that PCA analysis of human movements leads to conclusion that most of the variance is explained by few components as described in \cite{Sanger2000},
which confirms the old idea of the pioneer of kinesiology field of study Nikolai Bernstein. Bernstein proposed that human motion despite being high dimensional process can be described using points of low dimension.
In other words, motion primitives can be represented in a compact manner using small finite
number of variables for encoding. These variables are responsible to flexible adaptations to variations of related features such as positions of target objects, size and shape of objects, initial position of a manipulator, etc.

Motion primitives is an indispensable idea both in robotic and human behaviour modeling. 
They provide modularity in construction of complex interactions with an environment. Primitives are considered as a minimal set of reusable patterns to be combined to generate diverse patterns. 
The importance of motion primitive for robotic behaviour design is discussed in \cite{schaal1999}. 
In their work authors specify two types of motion primitives: discrete and cyclic, which correspond to a fixed point and a limit cycle in dynamic systems, respectively. 
And each of them can be represented by a single point in a parameter space of dynamic systems.

There are many approaches to learning behavioural or motion primitives. 
One of them is described in \cite{schaal2004}: direct modelling differential equations for discrete and oscillating patterns with variable parameters tuned by reinforcement learning.
Similar approach is taken in \cite{ude2010} with the emphasis on two types of primitives.
However the training is goal-oriented in both cases. 
The reward function is designed to ensure specific dynamic properties of a trajectory and ensure reaching certain final state. 
In reality, however, there are many constraints regarding interaction with objects in an environment in certain way which are hard to take into account when hand-designing reward functions are used.
It is theoretically possible to extract those constraints automatically via
supervised learning through imitation of recorded trajectories, if the data is quite plentiful.
For example, in \cite{noda2014} autoencoder is used to create a generative model for multimodal primitives.

We consider supervised learning scenario where every motion has finite encoding and can be regenerated using this encoding and shared generative model implemented as a Recurrent Neural Network. 
In our previous work \cite{nikulin2020}
we demonstrated that explicit embedding of hypersurfaces corresponding to each motion primitive in shared latent space enhances inter-primitive generalization capacity. 
However, this approach requires to manually label learning data, which may be infeasible when dealing with large datasets. 
In this paper we address the issue of finding suitable latent representation of sensory-motor data which can be automatically clustered for each corresponding primitive.

First thing to notice is high correlation between motor and sensory (typically visual) information. 
Information contained in a sequence of joint angles of a robotic manipulator allows us to partially generate visual sequence.
For example, the manipulator is reaching for an object and then pushing it in some direction. Sequence of joint angles
in this movement provides information about location of the object at any moment of time. 
Therefore by having appropriate encoding of kinematics data we can reconstruct sensory information with some precision. 
From these consideration, we assume that Kinematics data can be used to initialize values of latent variables for each sample prior to learning weights of RNN for generative model.

Clustering motion primitives is an intricate problem. 
Motions corresponding to different primitives could be located very close
to each other in trajectory space. 
For example, let us consider cases of a robotic manipulator reaching for grasping and reaching for simply touching an object in the same location in space.
Grasping and touching is an example of different primitives. 
Specific motions within each primitive are determined by location of the object in space. Hence, each primitive is a low-dimensional manifold in trajectory space. 
We can exploit an assumption about structure of these manifolds for clustering. 
Namely, they are close to linear, or at least could be embedded in 
low-dimensional linear subspaces. 
Linear subspace clustering has been a prominent topic in recent years.

However, even if we drop the assumption about linearity, we can linearly project low-dimensional manifolds from
trajectory space into parameter space without any overlapping of primitives. In \cite{calinon2007} authors
use projection of trajectory data into dominant principal components for further probabilistic modeling.
Corollary of Whitney embedding theorem described in \cite{Sauer1991} tells us that \textit{almost all}
linear projections will have the required property. Thus, a random linear projection will suffice.
Moreover, we show that random projection is also robust enough if the parameter space has sufficient dimension.

There are two hypothesis we test: (i) motion primitives form close to linear manifolds in trajectory space
and (ii) random projection of motion data to latent space and consecutive learning the generative model keeps 
linear manifolds at a degree enough for subspace clustering, which ensures robustness and separation of 
encodings for different primitives. To test the first hypothesis we will analyse generated joint-angle trajectories
of humanoid robot obtained via mapping motion-capture data. For the second hypothesis we design artificial data with
plentiful amount of trajectories to test generalisation capacity of the model. Two types of generalisation 
are tested: intra- and inter-primitive generalisation. The former is the ability of the model to generate 
unseen samples from known primitives and the later is the ability to quickly learn new primitives.

\section{Background}
\subsection{Generative Models and Predictive Coding}
According to predictive coding theory, formulated in \cite{Rao1999} and \cite{friston2006}, 
behaviour of an agent (e.g. robotic manipulator) can be modeled as
constant generation of predicted sensory information $\mathbf{o}_t$ based on 
some changing internal state $\mathbf{d}_t$ every moment of time $t$.
Sensory information includes proprioception, which allows to generate motor commands for robot to satisfy 
predicted future positions of manipulators. We split information $\mathbf{o}_t$ into two parts:
proprioception $\mathbf{m}_t$, which we will refer to as "motor commands" for the sake of brevity, and 
the rest of the perception $\mathbf{s}_t$.
Predicted information is compared with the real one and internal state is corrected based on the error.
Internal state dynamics can be modeled as a deterministic process by RNN \cite{annabi2021}, 
for instance MTRNN \cite{ahmadi2017}, LSTM \cite{graves2014}, or as 
a stochastic process, for example PVRNN \cite{ahmadi2019}.
Correction of the entire internal state is computationally costly due to high dimension. Therefore, sequence of internal states
is encoded in either sequence of latent vectors $\mathbf{z}_t$ or a single latent vector $\mathbf{z}$.
The former is the case of PVRNN. In this study we concentrate on the later since we work with simple motion primitives, which
could be encoded as points of a finite-dimensional space. 

Overall dynamics of the model of interest is described in the following equations:
\begin{subequations}
\label{model_dynamics}
\begin{align}
\label{model_dynamics1}   \mathbf{d}_0 &= g(\mathbf{z}) \\ 
\label{model_dynamics2}   \mathbf{d}_{t+1} &= f(\mathbf{d}_t, \mathbf{z}) \\ 
\label{model_dynamics3}   \mathbf{s}_t &= s(\mathbf{d}_t) \\ 
\label{model_dynamics4}   \mathbf{m}_t &= m(\mathbf{d}_t)
\end{align}
\end{subequations}
Here $\mathbf{z}$ determines the initial state and also controls the state transition. A variable which 
controls the state transition is called \textit{Parametric Bias} and typically independent from initial state as in \cite{tani2003}.
In this study we unify all information which describes a sensory-motor sequence in a single vector $\mathbf{z}$.
Detailed description of the utilized model will be provided in further sections.

\subsection{Random Linear Projections}
An important corollary of the Whitney Embedding Theorem, 
described in \cite{Sauer1991}, states the following:

\begin{thm}\label{whitney_thm}
Let $A$ be a compact subset of $\mathbb{R}^k$, $\text{lower
boxdim}(A) = d$. If $q > 2d$, then almost every linear transformation of $\mathbb{R}^k$ to
$\mathbb{R}^q$ is one-to-one on A.
\end{thm}

Where $\text{lower boxdim}(A)$ is lower box dimension of $A$.
In particular, if $A$ is a smooth manifold compactly restricted in some finite volume, the lower box dimension
will coincide with box-counting dimension and regular manifold dimension. Moreover, if $A$ is a union of
smooth manifolds, then its box-counting dimension is equal to maximal dimension of a manifold in the union.
In other words, if some high-dimensional data lie close to a union of at most $d$-dimensional manifolds,
then $n > 2d$ random observations are enough to encode the data without ambiguity.

This is a side result of the paper. The main statement the authors are making is about general smooth maps 
instead of linear. The space of all smooth maps is infinite-dimensional and there is no Lebesgue measure on such a
space. Therefore, the authors propose their own definition of the term "almost all" in terms of prevalence. However this
definition is not required here and the term "almost all" is used in usual measure-theoretic way. The space of all linear maps
is finite-dimensional. We refer the interested reader to the original literature cited above.

Let $p$ be number of variables in a single motor command for a robotic manipulator (e.g. joint angles values).
We regard a sequence of $T$ motor commands $\mathbf{m}_{1:T}$ corresponding to a single motion as a point in $\mathbb{R}^{pT}$.
The main assumption of this paper is that specific motions in motion primitive are specified by small number of variables.
For example, all motions corresponding to robotic manipulator touching an object could be encoded just in the
position of the touch point given the initial position of the manipulator is fixed. Meaning all these motions are
points on a smooth low-dimensional manifold in $\mathbb{R}^{pT}$. And all possible motions are described as points
on a union of such primitives. Since values for motor commands are restricted by their nature we can ensure all available
points are bounded in a finite volume, thus compact.
The above theorem guarantees that random linear transformation of $\mathbb{R}^{pT}$ to $\mathbb{R}^{q}$,
where $q$ is the dimension of latent representation, is one-to-one for all available motions given $q$ is sufficiently high.

\subsection{Linear Subspace Clustering}
Embedding theorem from the previous section guarantees the existence of an encoding, but it does not
provide any assurance about structure of the encoded information. In this section we explore one
particular assumption regarding organization of motor data. Namely, manifolds corresponding to motion primitives are 
close to linear. This enables us to apply \textit{Subspace Clustering} algorithm described in \cite{Vidal2014}.

The problem of subspace clustering is stated as follows: given a data matrix $Z = [\mathbf{z}_1, \mathbf{z}_2, \dots, \mathbf{z}_n]$
find correct labels $l(i)$ for each $\mathbf{z}_i$ under assumption $\mathbf{z}_i = \Phi_{l(i)}\mathbf{\xi}_i + \mathbf{\epsilon}_i$,
where $l(i) \in \{1, 2, \dots, K\}$ and $\Phi_{l(i)}$ are some linear projections from low-dimensional spaces.
In other words, we assume that data points $\mathbf{z}_i$ belong to the union of $K$ low-dimensional subspaces with errors $\mathbf{\epsilon}_i$.

\subsubsection{$\epsilon_i = 0$}
We start the investigation of this problem with simplified case when $\mathbf{z}_i$ belong to subspaces with no error.
There are many approaches to solve this problem. In this paper we concentrate our attention on self-representation matrix based ones.
Matrix $C$ is called \textit{self-representation matrix} when $Z = ZC$.
In other words, if we can represent each point $\mathbf{z}_i$ as a linear combination of all points,
matrix $C$ will be a matrix of linear coefficients. Consider skinny SVD decomposition of $Z$:
\begin{align}
    Z = U \Lambda V^T
\end{align}
Where $V$ is $r \times n$ matrix and $r$ is the rank of $Z$. Then matrix $Q = V V^T$ is a self-representation matrix:
\begin{align}
    Z V V^T = U \Lambda V^T V V^T = U \Lambda V^T = Z
\end{align}

\textit{Subspace Separation Theorem} formulated in \cite{Kanatani2001} states the following:

\begin{thm}
If the $\alpha$th and $\beta$th points belong to different subspaces, then $Q_{\alpha \beta} = 0$.
\end{thm}

We can interpret $Q$ as an adjacency matrix of a graph with nonzero entries indicating edges.
The intuition is the following: a collection of points belonging to the same linear subspace can be 
expressed as a linear combination of each other and have linear coefficients corresponding to the rest of points equal zero.
Then we can apply well-known spectral clustering algorithm based on SVD decomposition of Laplacian of $Q$ for finding desired labels.

\subsubsection{$\epsilon_i \neq 0$}
There are many techniques to approach fluctuations from the ideal case.
For example we can formulate the noiseless scenario as an optimization problem
with exact constraints and then relax them. Matrix $Q = VV^T$ is a solution to the following problem:
\begin{align}
    \min_C{rank(C)} \quad \textrm{s.t.} \quad Z = ZC \label{base_cluster_problem}
\end{align}
Indeed, $rank(C) \ge rank(ZC)$ and $Z = ZC$, thus $rank(C) \ge rank(Z)$.
We also know that $rank(VV^T) = rank(Z)$ by construction. Therefore,
matrix $Q$ defined above is a solution to the problem. However it is not unique.
In \cite{Liu2011} authors show that this is also the solution to convex analogue
of rank minimization problem:
\begin{align}
    \min_C{\|C\|_*} \quad \textrm{s.t.} \quad Z = ZC
\end{align}
Where $\|C\|_*$ is nuclear norm of matrix $C$. Since this variation of optimization
objective is convex, the solution is unique. This is the formulation of the 
optimization objective where, as proposed in \cite{Vidal2014}, we can
relax the constraints:
\begin{align}\label{relaxed_optim}
    \min_C{\|C\|_* + \frac{1}{\tau}\|Z - ZC\|_F^2} \quad \textrm{s.t.} \quad C = C^T
\end{align}
With $\tau$ being some hyperparameter. In this formulation we are allowed
to search for a matrix $C$ which is not exactly self-representation matrix, 
but close to it within some margin. The authors of this formulation
also prove that minimizer to (\ref{relaxed_optim}) is unique and
can be found in closed form. The optimal solution $Q^*$ is given by the
following formula:
\begin{align}\label{self_expression}
    Q^* = V P(\Lambda) V^T
\end{align}
Where operator $P$ acts on diagonal entries of $\Lambda$ as
\begin{align}
    P(x) = \begin{cases}
    1 - \frac{1}{\tau x^2} \quad &x > \frac{1}{\sqrt{\tau}}\\
    0 &x \leq \frac{1}{\sqrt{\tau}}
    \end{cases}
\end{align}
The topic of subspace clustering is full of intricacies. Number of proposed approaches
is growing every year. In this background we barely scratched the surface and covered
only those formulas which we utilized in our work.

\subsection{Affine Subspace Clustering}
The assumption that the subspaces are linear is too restrictive since it demands these subspaces to pass through the origin.
On the other hand, affine subspaces allow to have bias. Instead of expressing $\mathbf{z}$ as a linear combination of
each other, express them as an affine combination and solve the same problem of finding sparse representation of coefficient matrix.
Affine combination differs from linear by additional restriction to linear coefficients: they must sum up to one.
So problem (\ref{base_cluster_problem}) will be rewritten as
\begin{align}
    \min_C{rank(C)} \quad \textrm{s.t.} \quad Z = ZC, \mathbf{1}^T C = \mathbf{1}^T \label{modified_cluster_problem}
\end{align}
In fact, we can combine both restrictions in this optimization problem into one by constructing the following matrix
\begin{align}
    \tilde{Z} = 
    \begin{bmatrix}
    \mathbf{z}_1 & \mathbf{z}_2 & \dots & \mathbf{z}_n \\
    1 & 1 & \dots & 1
    \end{bmatrix}
\end{align}
This is a simple switch to homogeneous coordinates of $\mathbf{z}$.
Then, problem (\ref{modified_cluster_problem}) will have exactly the same form as problem (\ref{base_cluster_problem}):
\begin{align}
    \min_C{rank(C)} \quad \textrm{s.t.} \quad \tilde{Z} = \tilde{Z}C
\end{align}
Everything discussed for problem (\ref{base_cluster_problem}) also applies in this case.


\section{Problem Statement}

We define a single motion $\mathbf{o}$ as a sequence of sensor and motor pairs $\mathbf{o} = \{(\mathbf{s}_t, \mathbf{m}_t)\}$ of some fixed length $T$.
Given a set of observed motions $\{\mathbf{o}_i\}$ indexed by $i$, find low-dimensional latent vectors $\{\mathbf{z}_i\}$ together with
generative decoder $d$ such that $d(\mathbf{z}_i) = \mathbf{o}_i$ for all $i$. Moreover, it should be possible to find
motion clusters by analizing latent vectors $\{\mathbf{z}_i\}$ in unsupervised manner.

\section{Main Results}

To solve the given problem, we employed MTRNN architecture with second-order vertical connections and parametric bias.

\subsection{RNN Architecture}

\begin{figure}
    \centering
    \includegraphics[width=0.45\textwidth]{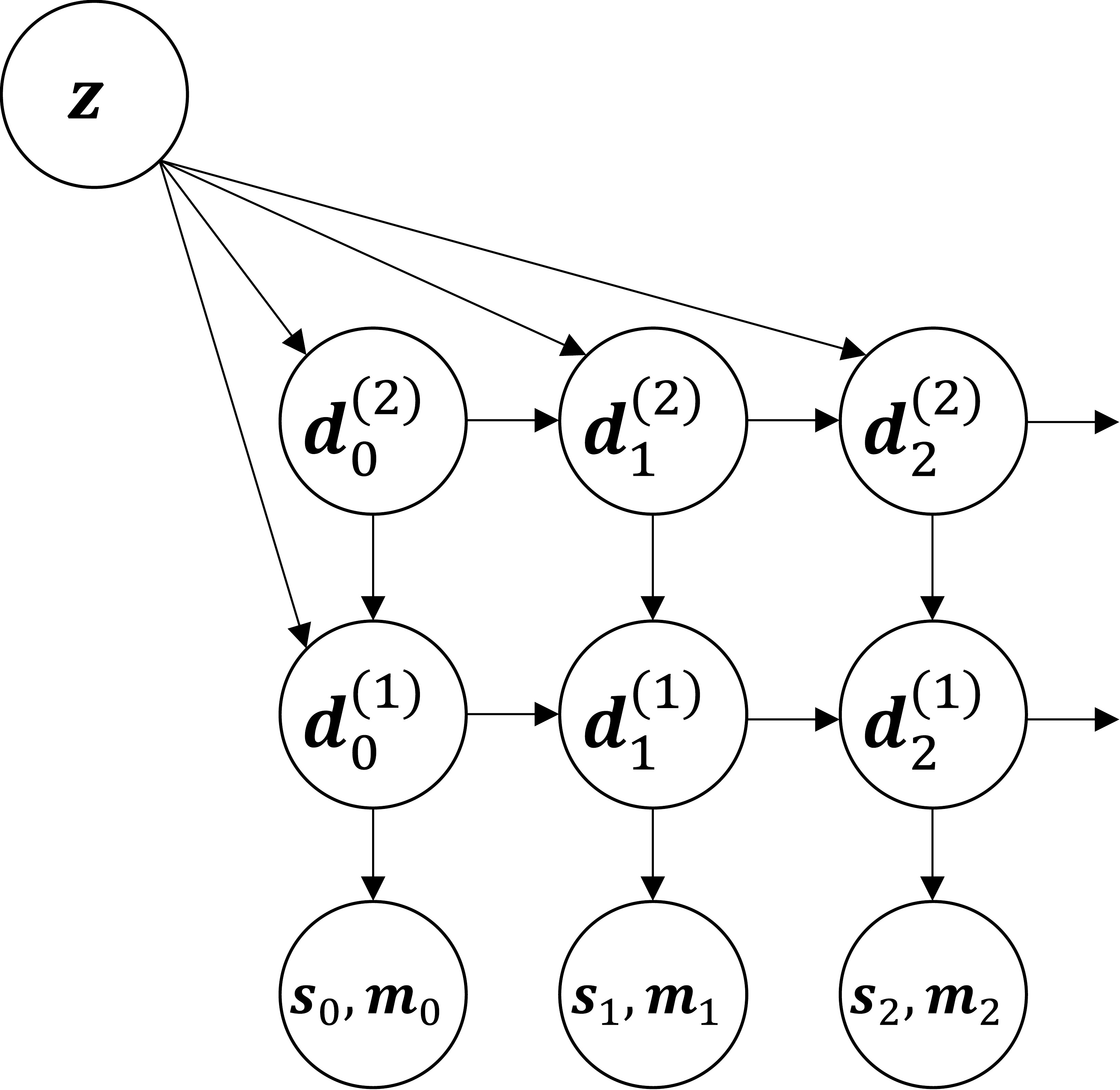}
    \caption{First three steps of unfolded recurrent schematic of the RNN architecture used in this paper. 
    There are two layers $\mathbf{d}_t^{(1)}$ and $\mathbf{d}_t^{(2)}$ in this illustration.}
    \label{fig:rnn_architecture}
\end{figure}

The base model is a variation of MTRNN architecture, see Fig. \ref{fig:rnn_architecture}. The state of the entire system at time $t$ is
encoded in a group of vectors $\{\mathbf{d}_t^{(j)}\}$ parametrised by $j$. Each vector $\mathbf{d}_t^{(j)}$ together with its
update function is referred as $j$th layer.
Every timestep each vector is updated according to the following formula:
\begin{equation}
\begin{aligned}
    \mathbf{h}_t^{(i)} = & \left(1 - \frac{1}{\tau}\right) \mathbf{d}_{t-1}^{(i)} \\
    & + \frac{1}{\tau} \left(W^{(i)} \mathbf{d}_{t-1}^{(i)} + U^{(i)} \mathbf{u}
    + \mathbf{A}^{(i)} \mathbf{d}_{t-1}^{(i)} \mathbf{u} + \mathbf{b}^{(i)} \right)
\end{aligned}
\end{equation}
\begin{align}\label{layer_norm}
    \mathbf{d}_t^{(i)} = LN(\mathbf{h}_t^{(i)})
\end{align}
Where $\tau$ is a timescale factor for a layer. Typically layers closer to the output have $\tau$ closer to one and are called \textit{fast layers}.
Conversely, layers further away from the output have higher $\tau$ and are called \textit{slow layers}. The idea is that the fast layers encode
more specific details about the current moment of the motion and slow layers encode more abstract information which doesn't change that fast.

Next, $\mathbf{u}$ is equal to vector from higher layer $\mathbf{d}_t^{(i+1)}$ for all layers but the very last. For the last layer $\mathbf{u}$
is equal to $p(\mathbf{z})$, where $p$ is some non-linear function (typically multi-layered perceptron). In such a way $\mathbf{z}$ plays the same
role for the last layer as any layer for the layer immediately below. Combining with the fact that higher layers change slower in time, 
we can interpret $p(\mathbf{z})$ as infinitely slow layer. It opens up opportunity to expand the model in the future and introduce dynamics
to $\mathbf{z}$. Perhaps combining motions one after another.

$W^{(i)}$ and $U^{(i)}$ are trainable weight matrices. $\mathbf{A}^{(i)}$ is trainable third-order tensor for multiplicative combination of
values of current and higher layer. Multiplication is very helpful compared to additive contribution. Remember that higher layer is very slow 
comparing to the current layer, it means simply adding linearly transformed values of upper layer would be almost as adding
a constant at each timestep, which, in turn, would facilitate diverging behaviour. $\mathbf{b}^{(i)}$ is also trainable bias vector.

$LN$ in (\ref{layer_norm}) is layer normalization.

Every timestep $t$ the output $(\mathbf{s}_t, \mathbf{m}_t)$ is generated as a non-linear transformation of the fastest layer $\mathbf{d}_t^{(1)}$.
To generate the motor command $\mathbf{m}_t$ a simple multi-layer perceptron is used. The sensory output $\mathbf{s}_t$ is an image
in our experiments. To generate it we used a couple of deconvolution layers with non-linear activation function.

Dynamics of the model match equations (\ref{model_dynamics}) stated in the background section.

Let $\theta$ be the vector of all model's trainable parameters.
Denote the entire output of the model during first $T$ timesteps as $d(\theta, \mathbf{z})$. Then loss function will be
\begin{align}
    L(\theta, \mathbf{z}_1, \mathbf{z}_2, ..., \mathbf{z}_n) = \sum_{i=1}^n {\| d(\theta, \mathbf{z}_i) - \mathbf{o}_i\|^2}
\end{align}
It is minimized with usual batch gradient descend. Important thing to notice is latent encoding vectors $\mathbf{z}_i$ are also trainable
parameters we need to optimize. And if initialization of RNN weights is a well-studied topic, initialization of $\mathbf{z}_i$ requires 
some investigation.

\subsection{Initialization of Latent Vectors}

There are two standard approaches to initialise latent variables: (i) set $\mathbf{z}_i$ randomly (e.g. according to standard gaussian distribution)
and (ii) set $\mathbf{z}_i = \mathbf{0}$.

Random initialization is detrimental for clustering. It doesn't provide any guarantee about structure of latent space.
First of all, motions corresponding to different primitives might be located very close
to each other in trajectory space, hence there is no guarantee for distance-based clustering results in latent space.
Then, manifolds corresponding to motion primitives are close to linear, according to our main assumption, in trajectory space.
But there is no assurance that projection of these manifolds to latent space after learning with random initialization will stay
close to linear. Non-linear manifold clustering is far more complicated problem.

Zero initialization leads to another problem. Because of batch nature of optimization algorithm,
every gradient step updates the entire $\theta$, but only fraction of $\{\mathbf{z}_i\}$. These irregularities in latent update
lead to faster convergence of $\theta$ compared to $\mathbf{z}_i$. And for reconstruction of the observed data
it is sufficient for $\mathbf{z}_i$ to be distinct enough. Hence, it is not to be expected for $\mathbf{z}_i$ to go far 
from initial point. The spread of latent vectors is close to zero, it is too unstable for clustering.

We propose novel way for initialization. Using results of theorem \ref{whitney_thm}, we can guarantee that random linear projection 
of the entire set of all observed motor sequences $\mathbf{m}_{1:T}$ into $\mathbf{z}$ will be one-to-one:
\begin{align}
    \mathbf{z} = P\mathbf{m}_{1:T}
\end{align}
Where $\mathbf{m}_{1:T}$ is a sequence of motor commands flattened into a single vector. $P$ is a random matrix of 
compatible dimension with $P_{ij} \sim \mathcal{N}(0, 1/q)$, where $q$ is the dimension of $\mathbf{z}$. With sufficiently expressive
generative model architecture it is possible to reconstruct these motor sequences based on the projections exactly:
\begin{align}
    \exists \theta : \mathbf{m}_{1:T} = d_m(\theta, \mathbf{z})
\end{align}
For sensory modality $\mathbf{s}_{1:T}$, it is highly correlated with motor modality, hence it requires minimal
correction to $\mathbf{z}_i$ to encode it properly.

Moreover, under the assumption of motion primitives being close to linear subspaces in trajectory space, the 
projected primitives are also close to linear subspaces in latent space if latent dimension suffice.
Detailed analysis of required dimensions to preserve subspace clusters is given in \cite{Li2018}.
Expected scalar product between any two projected vectors $P\mathbf{a}$ and $P\mathbf{b}$
are very close to the original product $\mathbf{a} \cdot \mathbf{b}$. More formally, the following expressions hold:
\begin{align}
    \EX_P[P\mathbf{a} \cdot P\mathbf{b}] &= \mathbf{a} \cdot \mathbf{b} \label{exp_prod} \\
    Var_P[P\mathbf{a} \cdot P\mathbf{b}] &\leq \frac{3\|\mathbf{a}\|^2 \|\mathbf{b}\|^2 - (\mathbf{a} \cdot \mathbf{b})^2}{q} \label{var_prod}
\end{align}
See appendix A for derivation. 
From this result it looks like map $P$ is almost an isometry for decent values of $q$. 
It is not true since we are projecting down, but there is a link with \textit{restricted isometry}
in the case when we have finite number of points. More details are provided by \textit{Johnson–Lindenstrauss lemma}. Unfortunately, this
lemma requires $q$ to be much higher than we need for efficient encoding. But isometry is not strictly required, since portion of topological
information is also contained in non-linear decoder $d$. Equations (\ref{exp_prod}) and (\ref{var_prod}) simply guarantee robustness
of a random projection, directions close to orthogonal in trajectory space will most likely stay close to orthogonal after the projection even
for not so high $q$ compared to required in Johnson-Lindenstrauss lemma.

The only assumption we must take is that correction to latent encodings $\mathbf{z}_i$ to accommodate sensory information
will not disrupt linearity.

\section{Experimental Results}

We prepare two experiments to attest hypothesis stated in the introduction. 
First is to apply subspace clustering algorithm for joint angle sequences of
robotic arms generated with demonstration by human via motion capture. Second is to do the same with artificially generated data for robotic
manipulator interacting with an object, and since this data is much more plentiful, do training of the RNN generative model with train/test
split of the data and then apply subspace clustering algorithm for learned latent encodings.
Furthermore, in the second experiment we compare quality of \textit{intra-}primitive generalisation for different modes
of initialization of latent variables as well as \textit{inter-}primitive generalisation.
Generalisation is the ability of a trained model to encode new samples with little or no change to
its parameters. Intra-primitive generalisation is about the encoding of samples belonging to known primitives,
inter-primitive generalisation is the capability to encode entirely new primitives.

\subsection{Motion Capture Data}

\begin{figure}[!t]
\centering

\begin{minipage}{0.48\textwidth}
\centering
\includegraphics[width=0.6\textwidth]{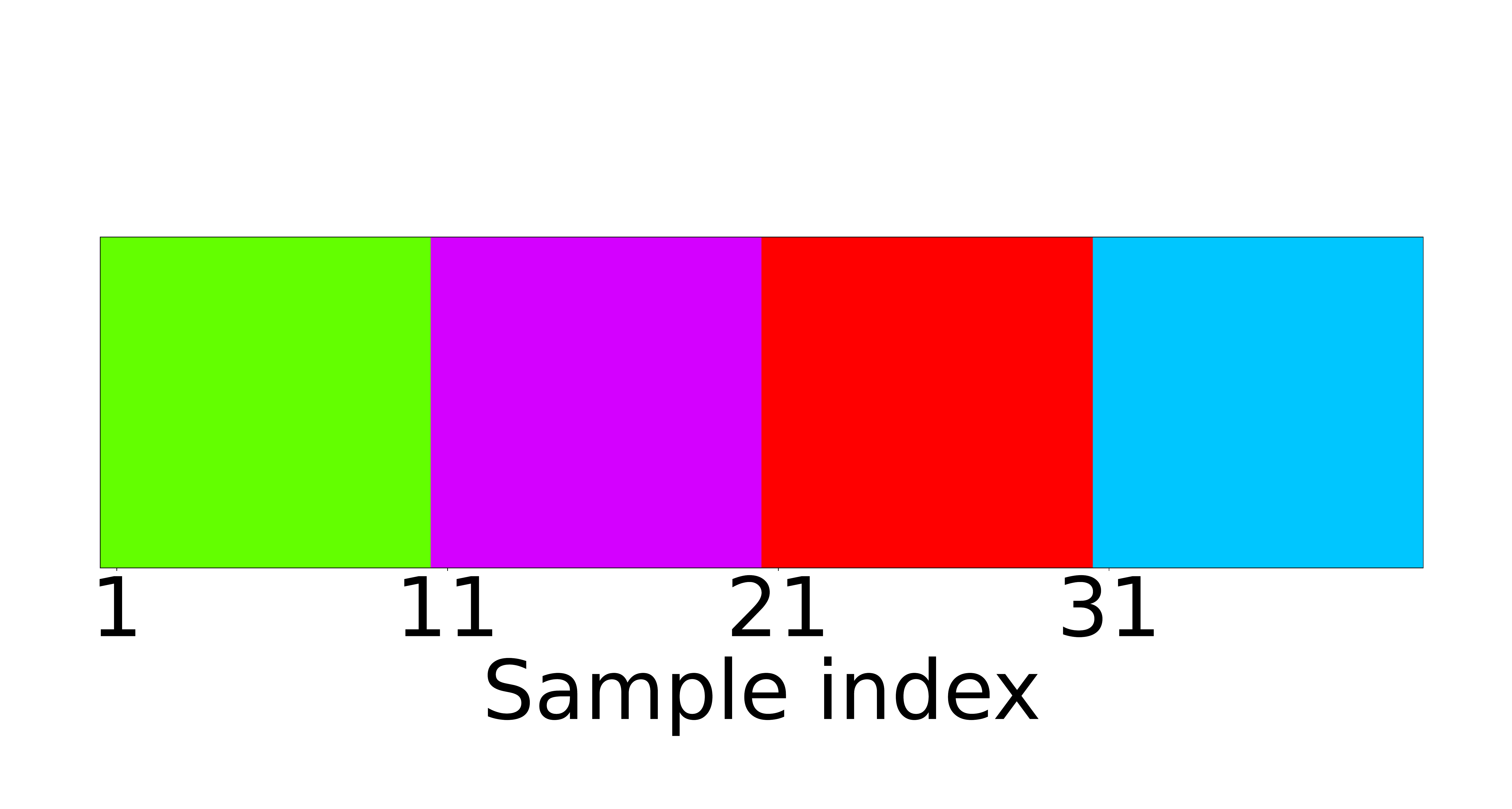}

(a) $\tau = 0.05$, $R^2 = 0.85$
\end{minipage}
\begin{minipage}{0.48\textwidth}
\centering
\includegraphics[width=0.6\textwidth]{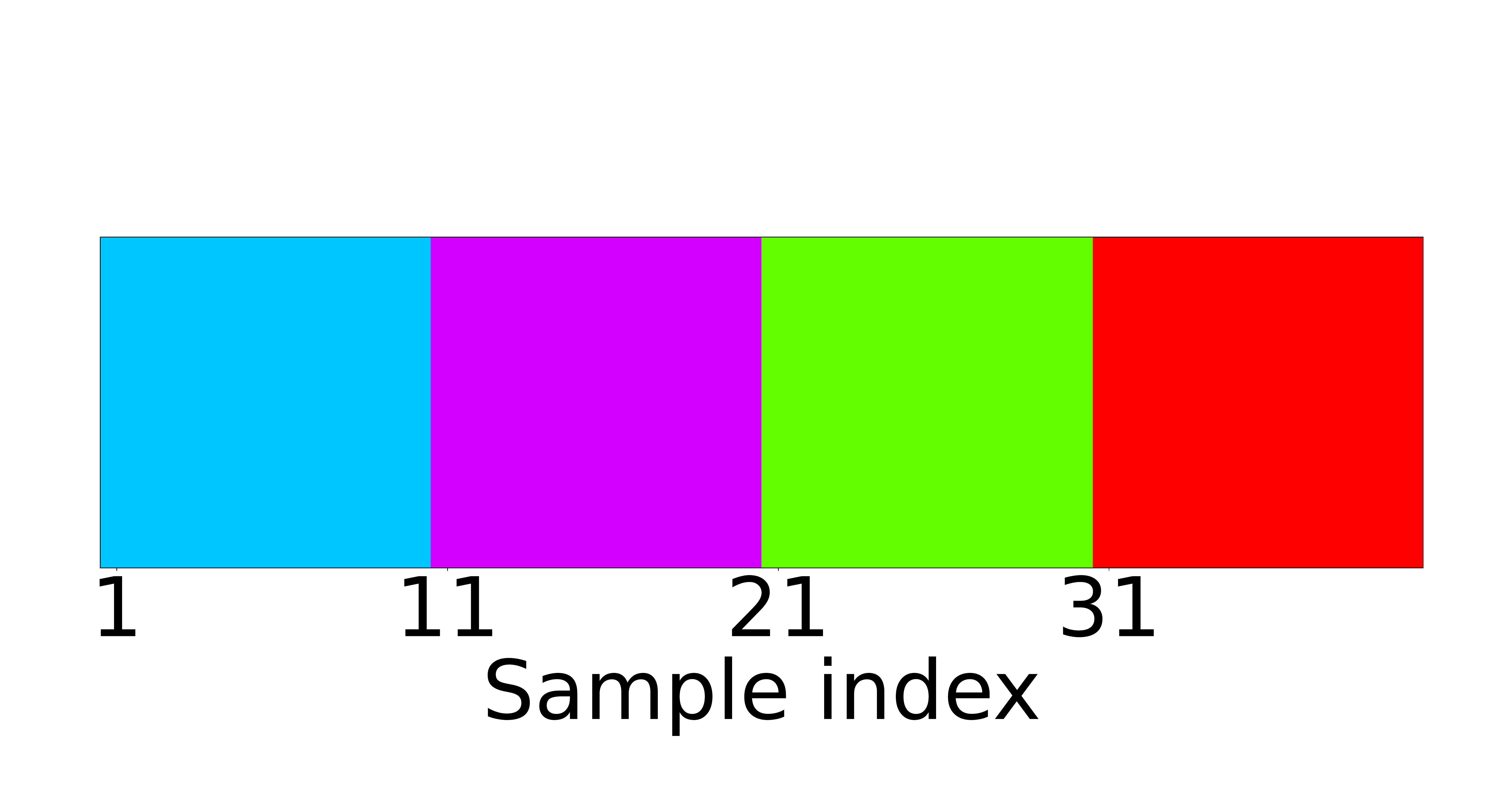}

(b) $\tau = 0.1$, $R^2 = 0.91$
\end{minipage}

\begin{minipage}{0.48\textwidth}
\centering
\includegraphics[width=0.6\textwidth]{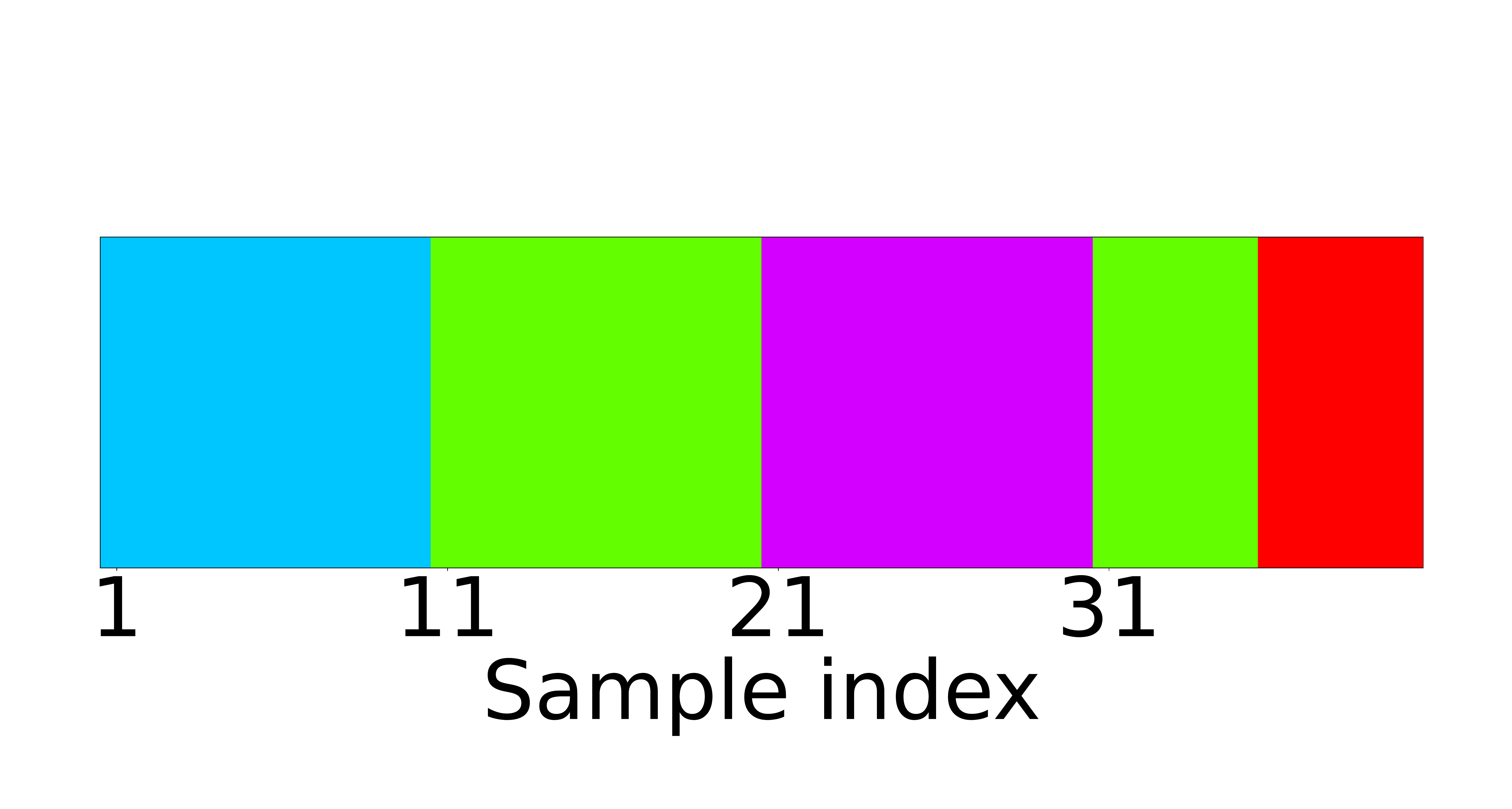}

(c) $\tau = 0.2$, $R^2 = 0.95$
\end{minipage}

\caption{Clustering results for robotic trajectory data acquired by motion capture. Different colors indicate different labels.
There are four ground truth clusters corresponding to sample indices 1-10, 11-20, 21-30 and 31-40. In cases (a) and (b) 
all labels identified correctly.}
\label{fig:torobo_clustering}
\end{figure}

In this experiment we use Torobo humanoid robot to generate motion trajectories. Torso and head are fixed, leaving only 12 joint angles to control
position of the two arms. We used motion capture device to manually control robot arms. Robot is interacting with a red cylinder object in front of it.
There are four motion patterns: (i) touch the object on top with left hand, (ii) touch the object on top with right hand, (iii) touch the object 
from the left with left hand and (iv) touch the object from the right with right hand. Each pattern is recorded 10 times with object being located
at different positions.

Recorded motions naturally vary in length. In order to align them to a single number $T$ of discrete timesteps, we used frequency domain
zero padding technique. Having a sequence $\mathbf{m}_{1:T_0}$ of joint angles with $T_0 < T$, compute the following:
\begin{align}
    \mathbf{f}_{1:T_0} &= \mathcal{F}(\mathbf{m}_{1:T_0}) \\
    \mathbf{\tilde{f}}_{1:T} &= [\mathbf{f}_{1:T_0/2}, \underbrace{0, 0, \dots, 0}_{T - T_0}, \mathbf{f}_{T_0/2:T_0}] \\
    \mathbf{\tilde{m}}_{1:T} &= \mathcal{F}^{-1}(\mathbf{\tilde{f}}_{1:T})
\end{align}
Where Discrete Fourier Transform $\mathcal{F}$ is applied to each joint angle sequence in $\mathbf{m}_{1:T_0}$ individually.
The resulting motion $\mathbf{\tilde{m}}_{1:T}$ will have the same "shape" as $\mathbf{m}_{1:T_0}$, but will have $T$ timesteps.

Furthermore, for each joint angle all the values in all sequences were normalized to be in [-1, 1] interval.
This was done to make contribution of each joint equally important for the trajectories.

We computed expression (\ref{self_expression}) for the acquired trajectory data with different values of $\tau$.
The choice of $\tau$ was made based on coefficient of determination $R^2$ for reconstruction of trajectories based on 
estimated self-expression matrix $Q^*$. For $0.85 < R^2 < 0.95$ the corresponding values of $\tau$ are $0.05 < \tau < 0.2$.
Then, using spectral clustering algorithm on obtained matrix $Q^*$ we put labels on each sample, see Fig. \ref{fig:torobo_clustering}.

The result shows that subspace clustering algorithm is able to correctly assign labels for different motion patterns in the case
of very noisy human made data. Which confirms our hypothesis about close to linear distribution of points belonging to 
the same motion primitive in trajectory space.

\subsection{Artificial Data}

\begin{figure}
\centering
\includegraphics[width=0.35\textwidth]{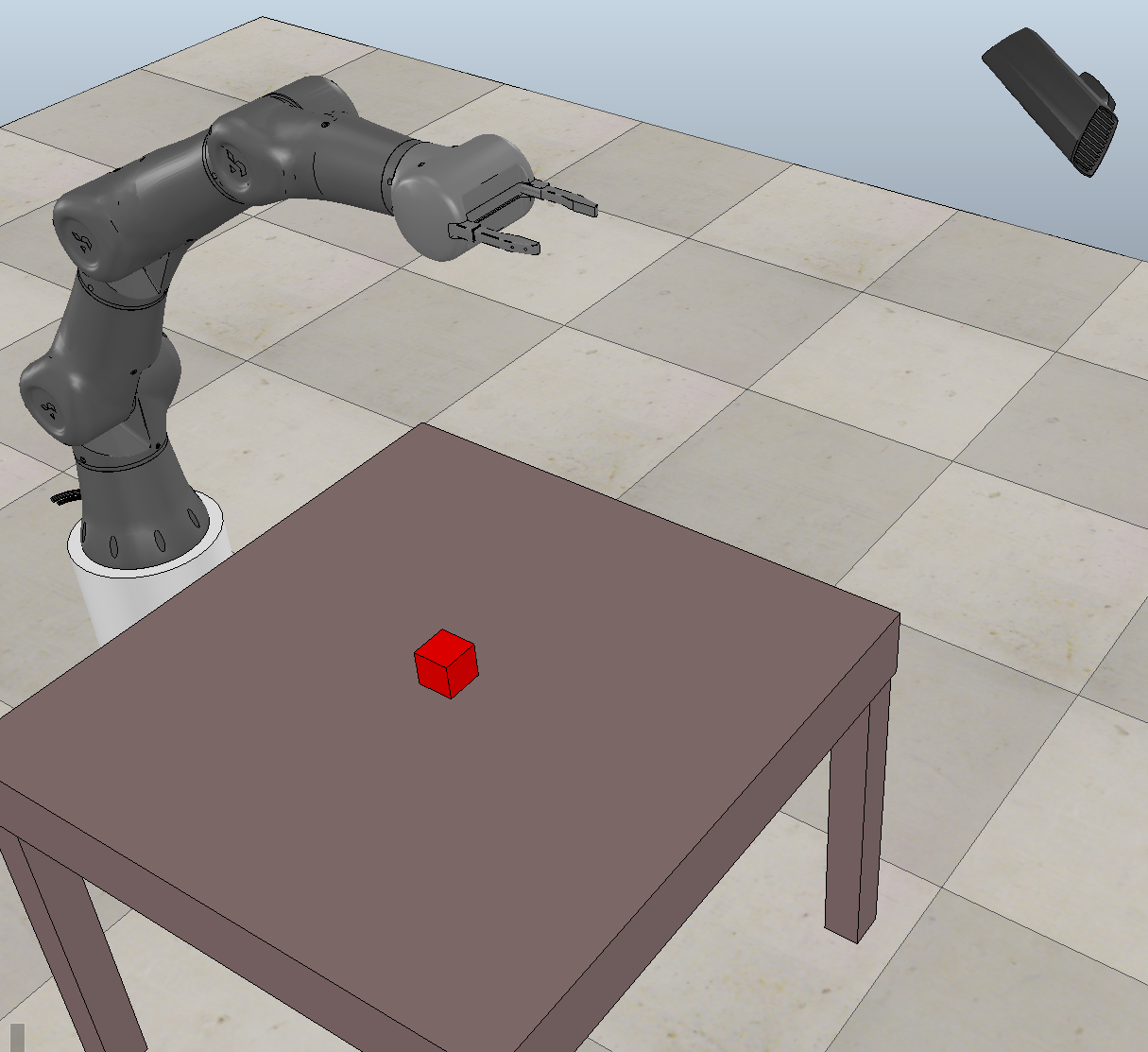}
\caption{Experimental setup includes robotic manipulator, red block on the table and the camera capturing RGB images.}
\label{fig:robot_setup}

\end{figure}

In the second experiment we used CoppeliaSim simulator \cite{CoppeliaSim} to generate a wide range of trajectories
for Torobo Arm manipulator. Torobo Arm manipulator position is encoded with 7 joint angle positions and 2 finger positions.
The experimental setup consists of the manipulator, table in front of it, object to interact with and a camera to 
record RGB images every timestep, see Fig. \ref{fig:robot_setup}.

We manually scripted the movement of the tip of the manipulator.
All the trajectories are acquired via inverse kinematics solver built in the simulator. Values of
joint angles across all motions are normalized to be bounded by the interval $[-1, 1]$.
There are 7 different motion primitives, each includes 245 motions for different positions of the block resulting in
1715 total samples. All motion primitives start with the manipulator approaching the block.
And then the behaviour for each primitive is the following:

\begin{itemize}
    \item Touch the block on top and stop.
    \item Grasp the block with two fingers and stop.
    \item Push the block to the furthest side of the table.
    \item Pull the block to the closes side of the table.
    \item Repeatedly touch the block from the right.
    \item Do circular motion around the block in clockwise direction.
    \item Do circular motion around the block in counter-clockwise direction.
\end{itemize}

Within each primitive the difference between specific motions is only in the position of the block on the table and size of the block.
Robotic arm always starts from the same position. So in trajectory space points belonging to each primitive
form 3-dimensional manifolds.

Every motion takes one minute of real time and is divided into 61 timesteps. At each timestep $t$, beside motor information, we also record
RGB image $\mathbf{s}_t$ of the size $48 \times 64$ pixels. The goal is to assign latent vector $\mathbf{z}$ to 
every pair of sequences $(\mathbf{m}_{1:T}, \mathbf{s}_{1:T})$ and build a decoder $d$ such that $d(\theta, \mathbf{z}) = (\mathbf{m}_{1:T}, \mathbf{s}_{1:T})$.
The decoder is RNN model from previous section. We used two dynamic layers, 32 variables in the fast layer and 12 in the slow layer.
We test three different ways to initialize latent variables $\mathbf{z}$ packed into a single matrix $Z$ 
described in the previous section: (i) zero initialization,
(ii) random initialization and (iii) random linear projection of the motor trajectory data.

\subsubsection{Intra-primitive generalisation}

To evaluate the model we split the entire dataset into two parts: one is used to train both $\mathbf{z}$ for each sample and model parameters $\theta$,
and another is to train $\mathbf{z}$ with parameters $\theta$ fixed. We refer to these
parts as "train dataset" and "evaluation dataset". Only 20\% of the entire dataset is randomly assigned to be train part, and the rest is for evaluation. Samples of all primitives are present in both train and evaluation datasets, so we indeed test intra-primitive generalisation.

\begin{figure}
\centering
\includegraphics[width=0.475\textwidth]{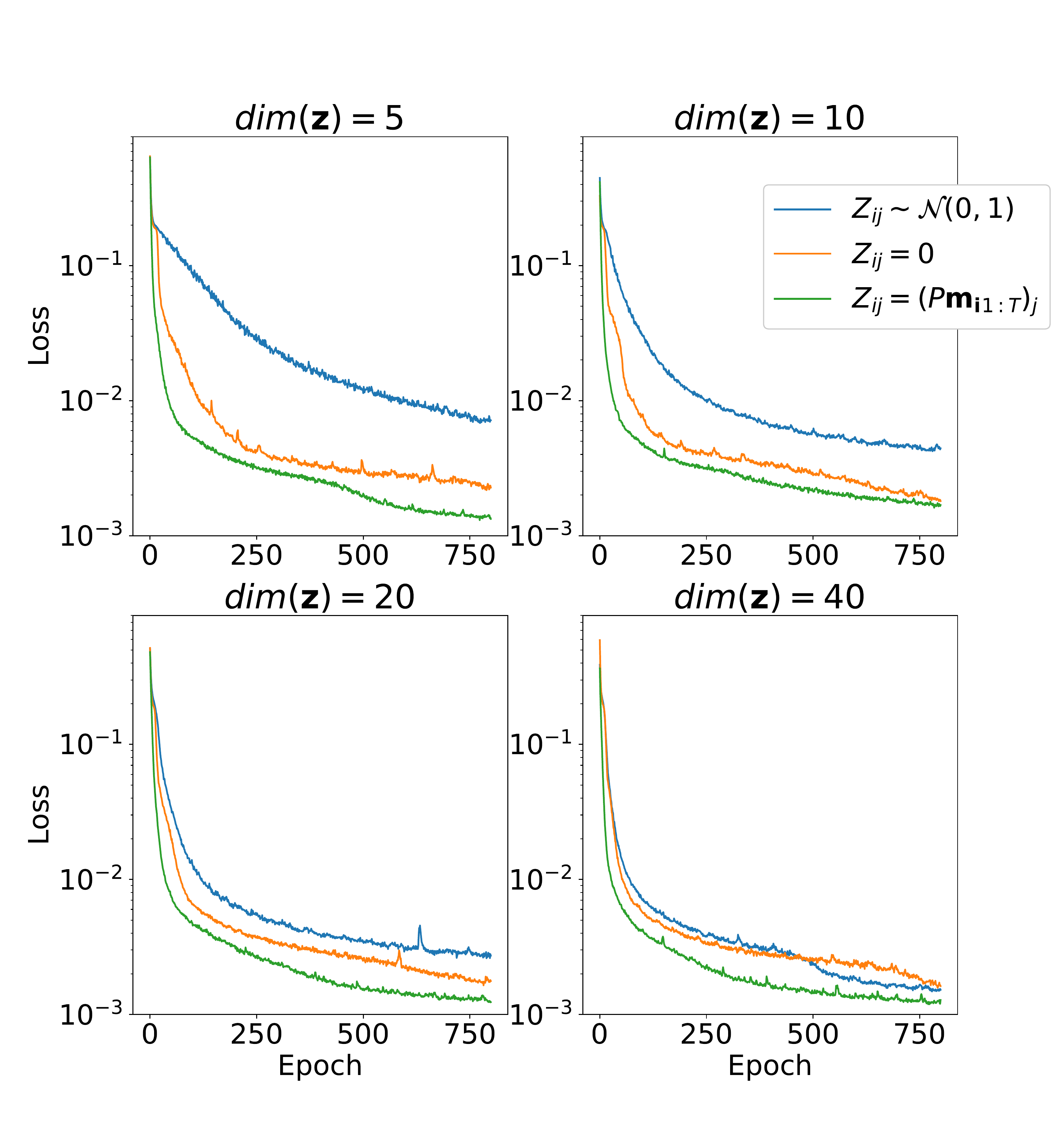}
\caption{Convergence of the loss function over 800 training epochs for different ways to initialize matrix $Z$ and different dimensions of $\mathbf{z}$. Both model parameters $\theta$ and matrix $Z$ are trained.}
\label{fig:loss_train}

\end{figure}

\begin{figure}
\centering
\includegraphics[width=0.475\textwidth]{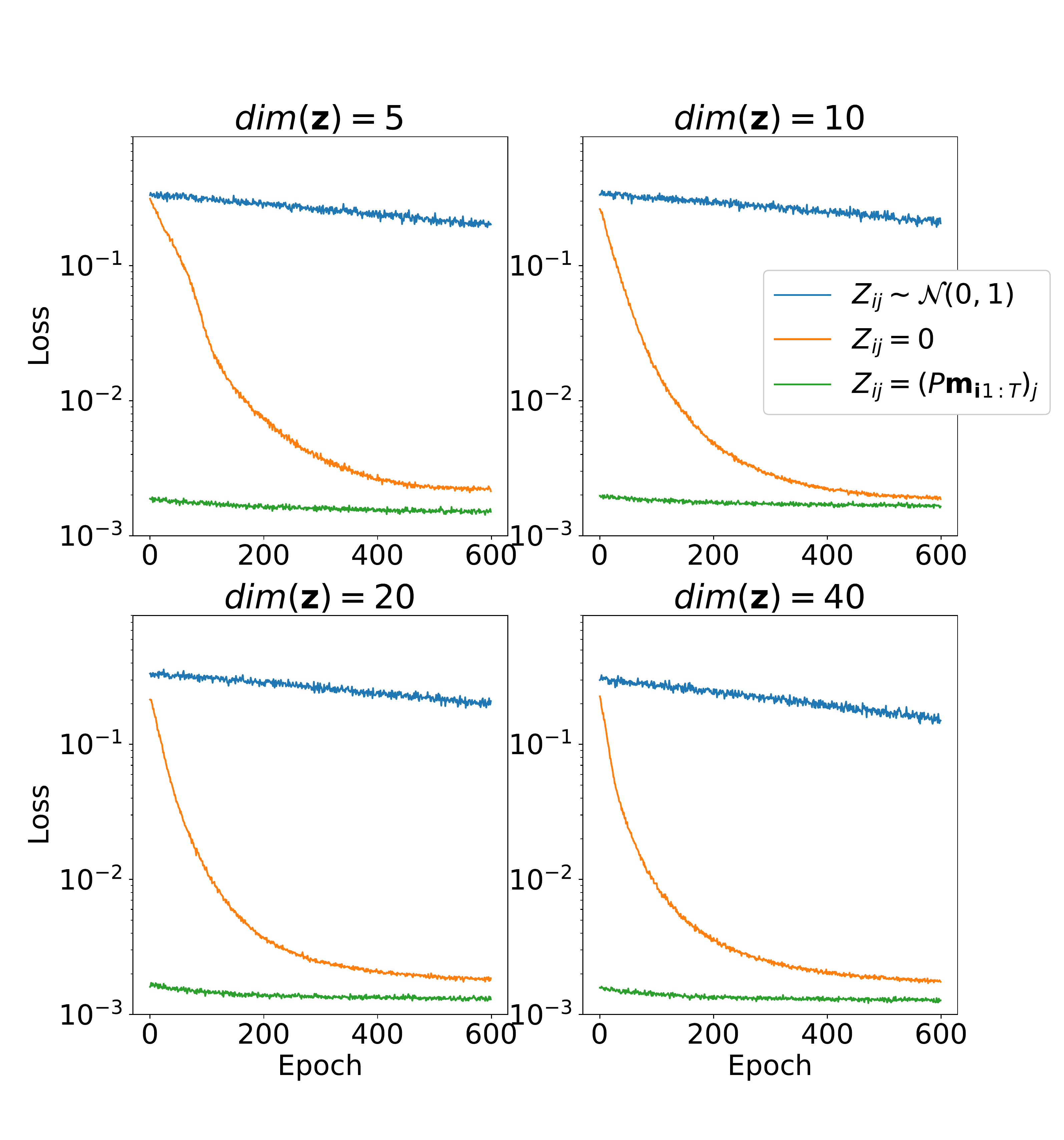}
\caption{Convergence of the loss function over 600 training epochs for different ways to initialize matrix $Z$ and different dimensions of $\mathbf{z}$. Only matrix $Z$ is trained, model parameters $\theta$ are fixed.}
\label{fig:loss_test}

\end{figure}

The results of training the model parameters $\theta$ together with latent variables $\mathbf{z}$ for training part
of the samples for different dimension of latent space are depicted in Figure \ref{fig:loss_train}.
As you can see, random initialization performs very badly for low-dimensional latent space.
Zero initialization, on the other hand, differs very slightly from random linear projection case.

The results of training only the latent variables $\mathbf{z}$ for evaluation part of the dataset with model parameters
$\theta$ being fixed are depicted in Fig. \ref{fig:loss_test}. Notice that the case with random initialization
did not converge at all. This means the model trained in this way has no generalization capacity and
is not capable to represent trajectories it did not see during the first training part. On the other hand,
initialization with random linear projection yields good results already without any training, which means
the initial distribution of projected points did not change much during the first part of the training. 
It was good from the start, and even addition of the visual information to the loss function didn't affect much.

It may look like resulting performance of trained models for zero latent initialization and random projection latent
initialization is very similar, but
some primitives from the experimental setup require more precision than others. In particular, pushing and pulling the block
as well as grasping is easy to fail if the gripper fingers miss just a bit. The success rate of these tusks determined by
qualitative assessment of recorded video is presented in Table \ref{assessment_table}.

\begin{table}
\centering
\caption{Success rates of trained models for sensitive primitives}
\label{assessment_table}
\begin{tabularx}{0.45\textwidth}{c|cc}
\hline
Primitive & $Z_{ij} = 0$ & $Z_{ij} = (\mathbf{m_i}_{1:T})_j$ \\
\hline
Pull the block & 0.86 & 0.92 \\
Push the block & 0.9 & 0.94 \\
Grasp the block & 0.8 & 1.0 \\
\hline
\end{tabularx}
\end{table}

Resulting generated joint angles and image sequences are very close to ground truth for all methods,
refer to Appendix B to see the comparison of generated motor and sensory data with ground truth.

\begin{figure}
\centering

\begin{minipage}{0.65\textwidth}
\centering
\includegraphics[width=0.5\textwidth]{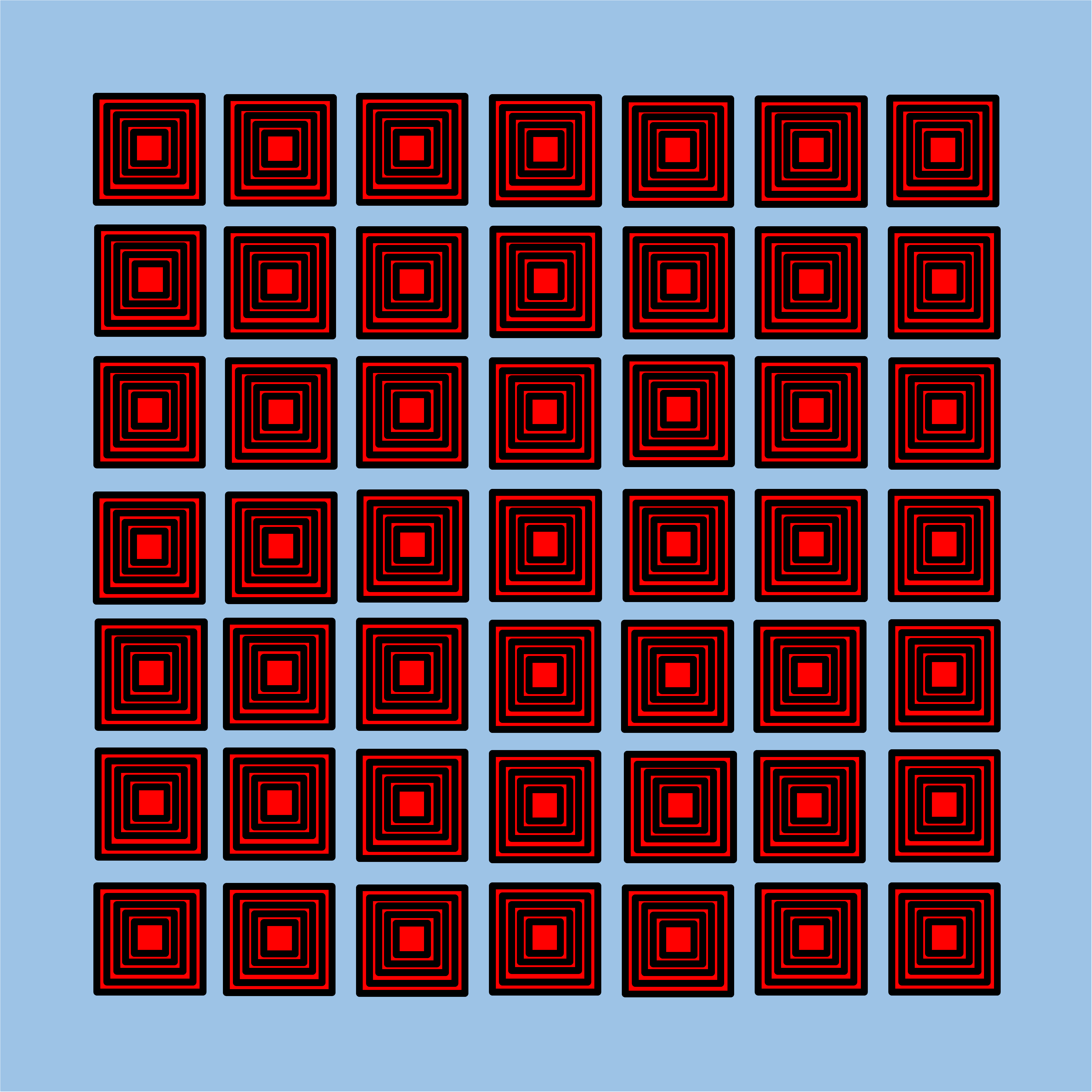}

Initial positions of the block on the table
\end{minipage}

\begin{minipage}{0.48\textwidth}
\centering
\includegraphics[width=0.65\textwidth]{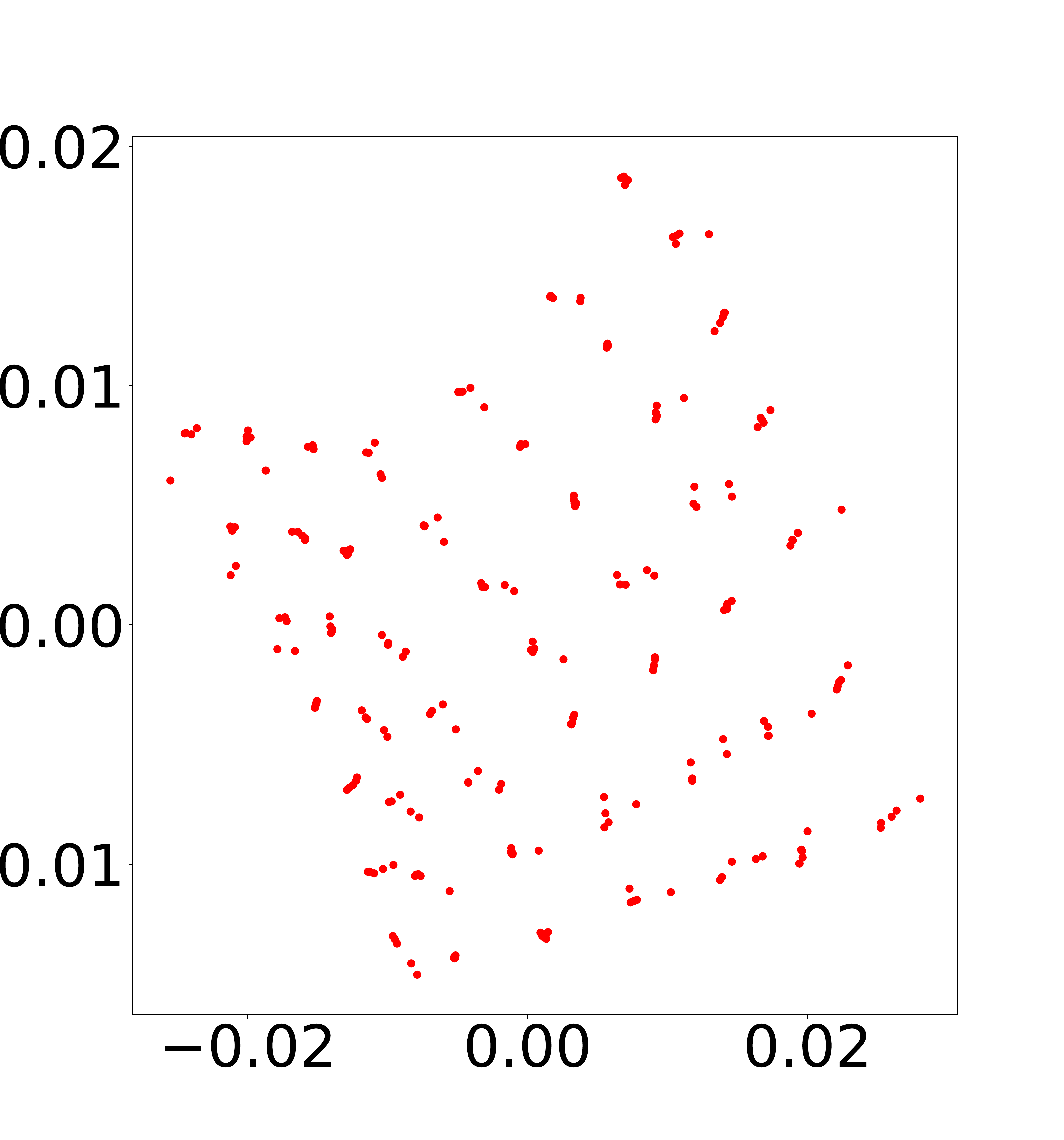}

$Z_{ij} = 0$
\end{minipage}
\begin{minipage}{0.48\textwidth}
\centering
\includegraphics[width=0.65\textwidth]{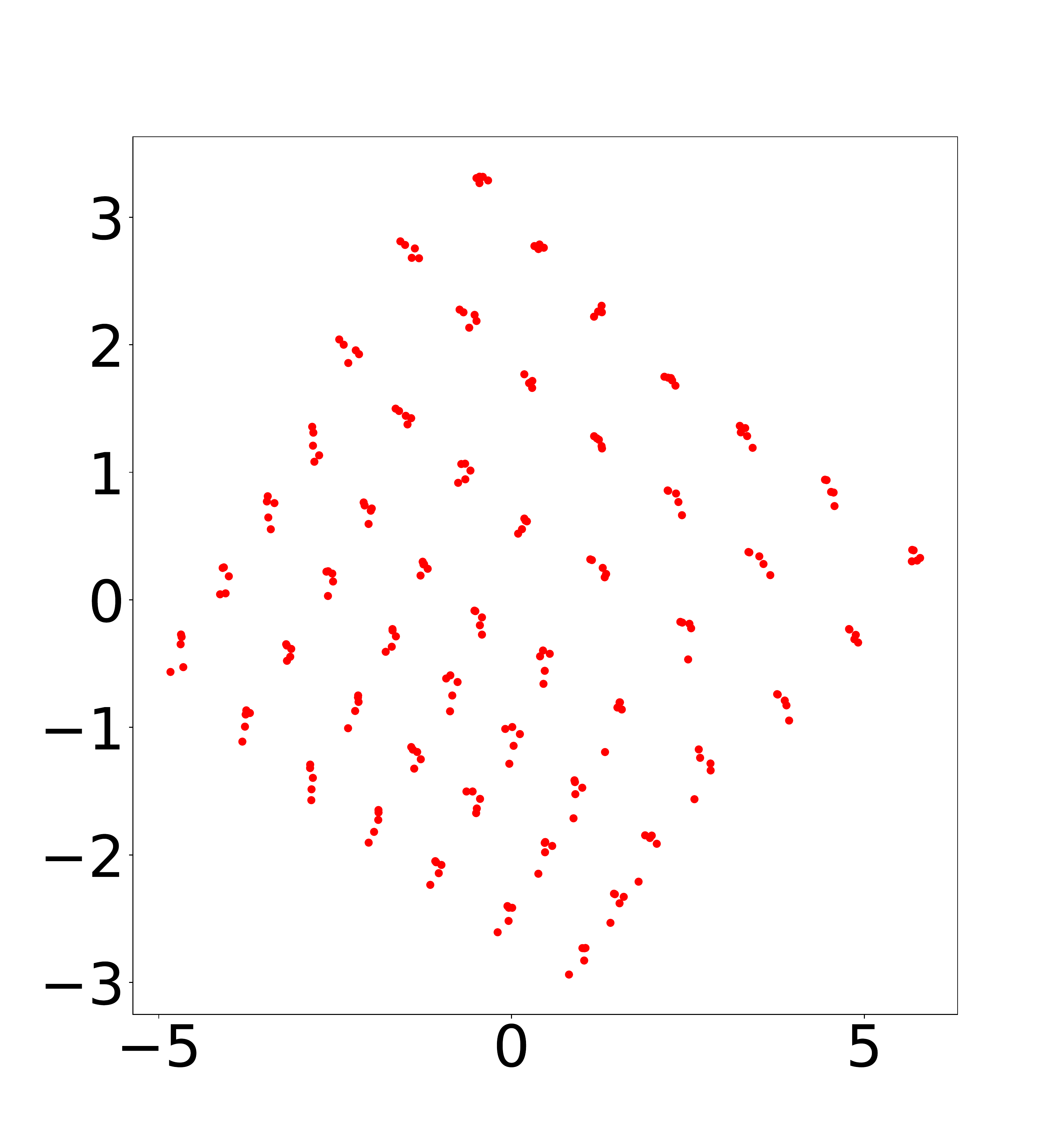}

$Z_{ij} = (P\mathbf{m_i}_{1:T})_j$
\end{minipage}

\caption{Initial positions of the block are depicted in (a), each position corresponds to five samples 
differing in block size.
First two principal components of the learned latent vectors for samples belonging to one specific primitive
are depicted in (b) for zero initialization method and in (c) for random linear projections initialization method.}
\label{fig:latent_encoding}
\end{figure}

By examining resulting latent vectors corresponding to samples from one primitive, we can see that 
random linear projection initialization method yields more "structured" result, see Fig. \ref{fig:latent_encoding}.
3-dimensional visualisation of first three principal components of latent vectors corresponding
to samples of three primitives is available at the following link: https://doi.org/10.6084/m9.figshare.19235034.v2

\subsubsection{Affine subspace clustering of latent space}

\begin{figure}[!t]
\centering
\begin{minipage}{0.9\textwidth}
\centering
\includegraphics[width=0.5\textwidth]{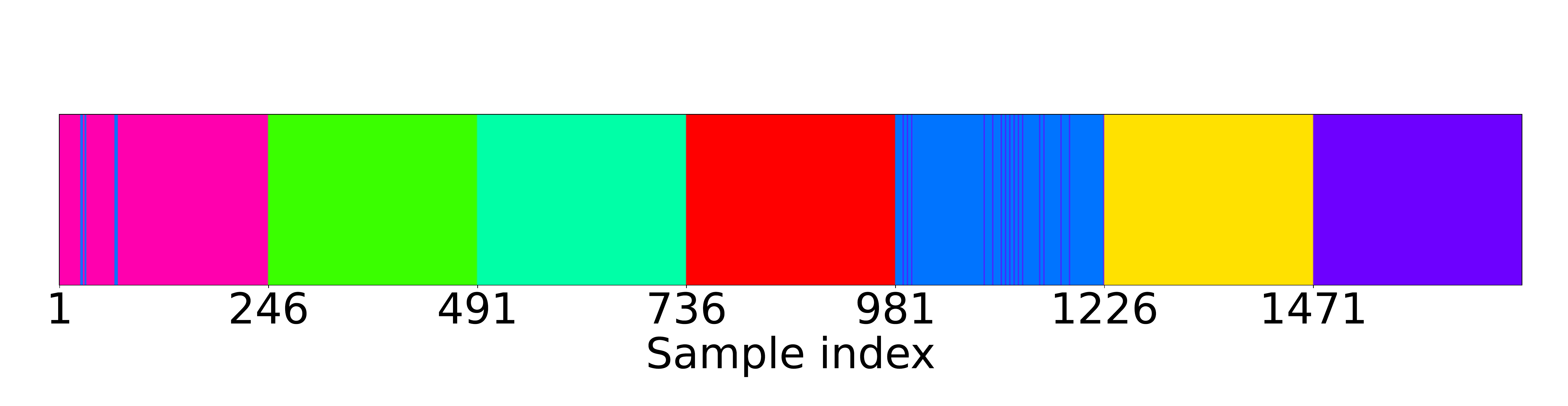}

$Z_{ij} = 0$
\end{minipage}

\begin{minipage}{0.9\textwidth}
\centering
\includegraphics[width=0.5\textwidth]{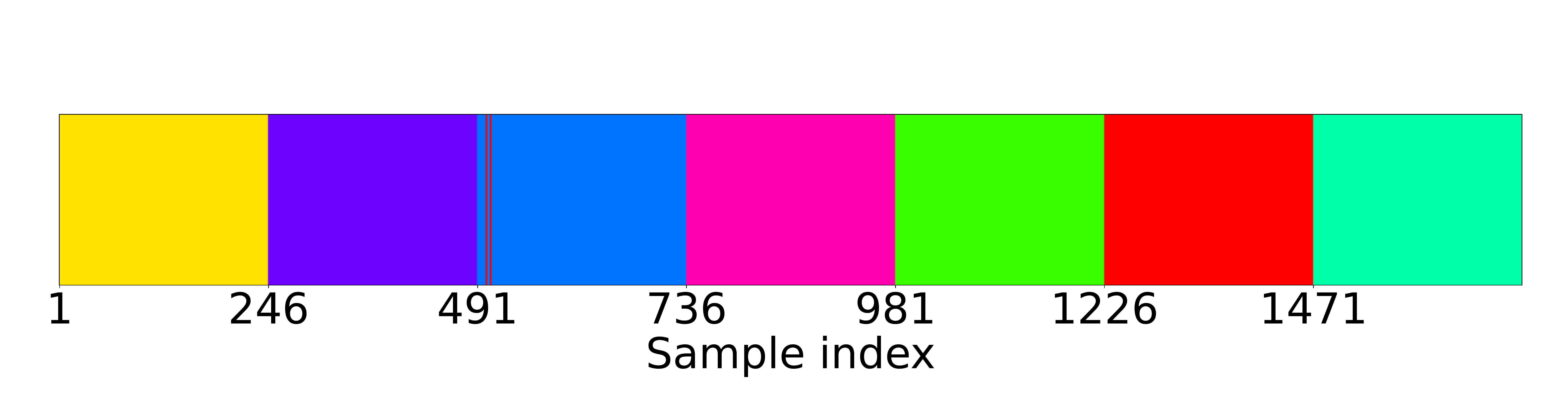}

$Z_{ij} = (P\mathbf{m_i}_{1:T})_j$
\end{minipage}

\caption{Clustering results for latent encodings for different initial values of $Z$. 
All 1715 samples are presented in order of consecutive 245-element batches
where every batch is a collection of motions belonging to the same primitive. Coefficient $\tau$ is set in such a way,
that in both cases coefficient of determination $R^2 = 0.9999$. It means resulting manifolds are very close to linear.}
\label{fig:arm_clustering}
\end{figure}

Next, to compare affine subspace clustering results we won't even consider the case of random initialization 
since it does not correctly encode evaluation part of the dataset. We use algorithm discussed in the background section
for matrix $Z$ to predict cluster labels for the data obtained by two steps of training and evaluation.
Comparison of the two methods to initialize $Z$ are depicted
in Fig. \ref{fig:arm_clustering}. Both cases are for $dim(\mathbf{z}) = 40$. They yield similar results,
meaning even without presenting motion information for initial values of latent variables, they self-organize
in a union of close to linear manifolds. It shows that initial values of latent variables obtained by 
proposed random linear projection already have some properties of fully trained latent representation.

\subsubsection{Inter-primitive generalisation}

\begin{figure}
    \centering
    \includegraphics[width=0.45\textwidth]{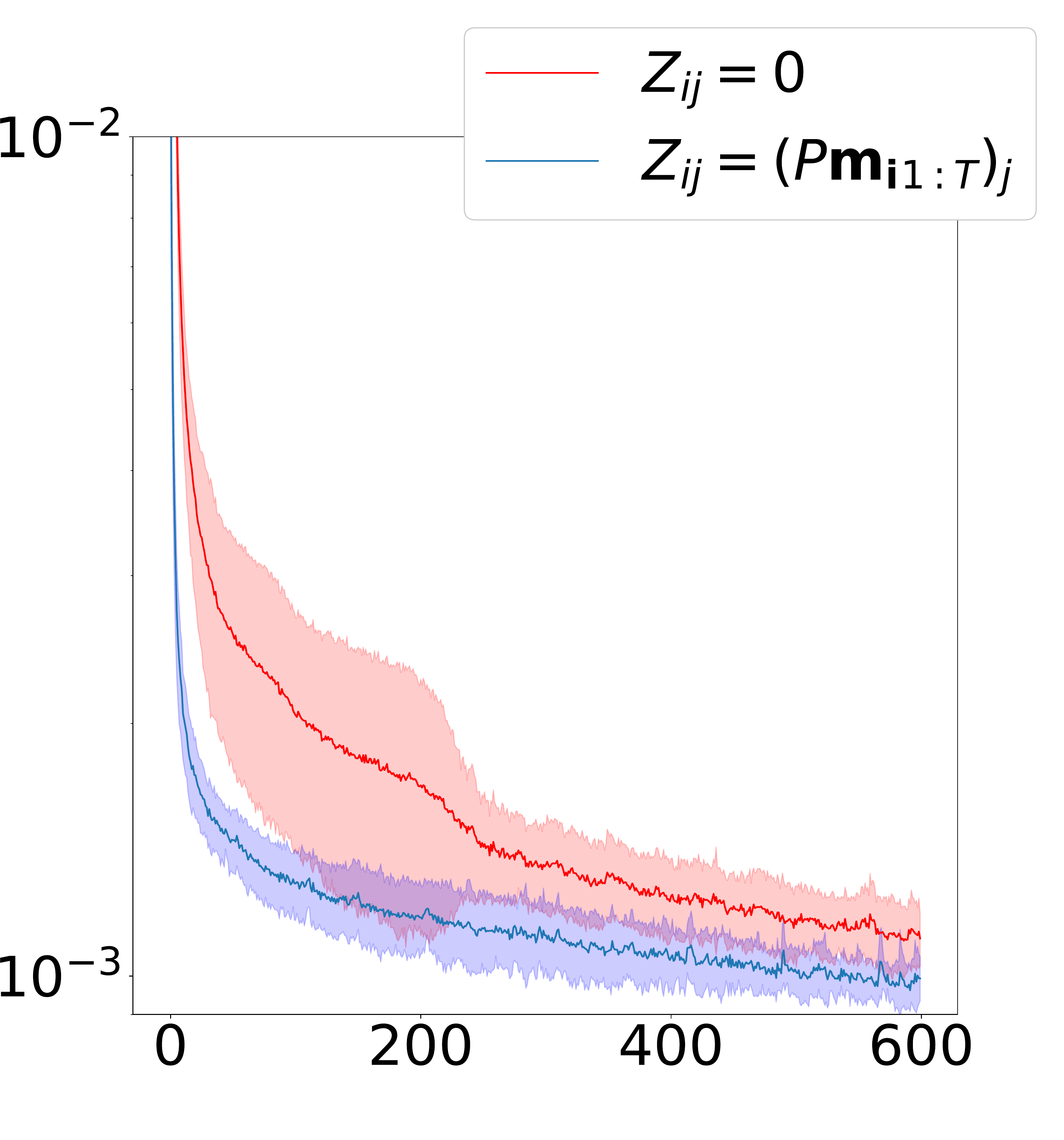}
    \caption{Mean and standard deviation of the convergence of the loss function in the second part of 
    inter-primitive generalisation experiment. The dimension of latent vectors is 20.}
    \label{fig:loss_inter}
\end{figure}

To test inter-primitive generalisation ability we perform seven independent tests for each primitive with different splits of 
the whole dataset into two parts: first part contains samples of six primitives for initial train of the model
and the second part contains the samples of the remaining primitive.
We also add a small portion (10\%) of samples from the first part to the second part to avoid forgetting.
Incremental learning is a difficult topic and we won't go into its intricacies by preferring simple solution.
Both model parameters $\theta$ and latent encodings are trained in both parts, but parts are trained consecutively.

Since this problem is considerably harder than intra-primitive generalisation, we compare only two successful
latent variables initialisation methods from the previous experiment: zero initialisation and proposed random linear projection.
Results of the training the second part are depicted in Fig. \ref{fig:loss_inter}. Proposed method shows slightly improved
results in terms of convergence speed, final value of the loss function and robustness of the learning.

\section{Summary and Discussion}
In this paper we investigated structure of robotic motion primitives in trajectory space and
ways of efficient encoding of that structure. Distinctive feature of each primitive is that the set of 
all motions belonging to this primitive lies on a low-dimensional manifold embedded in trajectory space,
which was confirmed by experiment in which we were able to reconstruct artificially generated robotic motions
from random linear projection of its motor trajectory data using RNN model. 
Moreover, these manifolds are close enough to affine subspaces, which enables us to use subspace clustering algorithm for
labeling a collection of motion in unsupervised manner.
This claim comports with clustering results of data obtained by motion-capture device.
Another assumption is correlation of visual information with motor commands. In our experiments we showed
that only slight correction to initial values of latent variables obtained by random linear projection is 
required to minimize combined loss function for motor and visual data.
Last thing to show was random linear projections do not disturb affine subspace clusters of trajectory space,
it is still possible to do subspace clustering of projected data for sufficiently large dimension of latent space.
Clustering results for latent encodings show sufficient precision to support this claim.
Initialization of the latent variable by random linear projections improves intra- and inter-primitive 
generalization capabilities compared to conventional initialization methods.

To model more complex behaviours composed of many consecutive primitives, instead of
using a single latent vector $\mathbf{z}$ of fixed dimension usually a sequence of such vectors is used.
For the future research we are planning to extend latent variable initialization algorithm by random linear projections
to a sequence of latent vectors. It can be done via one-dimensional convolution through time of a random linear
projection with long motor sequence. There are some challenges such as primitives might have different number
of timesteps and not clear borderline between primitives.

Another point is visual perception information is not used to initialize encodings. The problem here is only a
small portion of each perceived image is relevant to motion, the rest is a background noise which will 
clutter random projection. Some attention mechanism will potentially alleviate the problem. Vision information is
highly correlated with motor commands, but there is some part of it which is independent, such as
colors of the objects the robot is interacting with.

\printbibliography

\section*{Appendix A: Random Projection Robustness}
$\mathbf{a}$, $\mathbf{b} \in \mathbb{R}^k$, $P$ is a matrix of compatible dimension which defines a random linear map from 
$\mathbb{R}^k$ to $\mathbb{R}^q$, $P_{ij} \sim \mathcal{N}(0, 1/q)$. The derivation of (\ref{exp_prod}) is straightforward:
\begin{align}
    \EX_P[P\mathbf{a} \cdot P\mathbf{b}] = \EX_P\left[\sum_{j=1}^q{\left(\sum_{i=1}^k{P_{ij} a_i}\right)\left(\sum_{i=1}^k{P_{ij} b_i}\right)}\right] \label{exp_q_begin}\\
    = \EX_P\left[ \sum_{j=1}^q \left( \sum_{i=1}^k{P_{ij}^2 a_i b_i} + \sum_{i=1}^k \sum_{l \neq i}^k{P_{ij} P_{lj} a_i b_l} \right) \right] \\
    = \sum_{j=1}^q \left( \sum_{i=1}^k{\underbrace{\EX_P[P_{ij}^2]}_{=1/q} a_i b_i} + \sum_{i=1}^k \sum_{l \neq i}^k{\underbrace{\EX_P[P_{ij}]}_{=0} \underbrace{\EX_P[P_{lj}]}_{=0} a_i b_l} \right) \\
    = \sum_{j=1}^q \left( \sum_{i=1}^k{\frac{1}{q} a_i b_i} \right)  \\
    = \frac{1}{q} \sum_{j=1}^q \left( \sum_{i=1}^k{ a_i b_i} \right) \label{exp_q_end}\\
    = \mathbf{a} \cdot \mathbf{b}
\end{align}
$j$th component of vector $P\mathbf{a}$ is a normally distributed random variable:
\begin{align}
    (P\mathbf{a})_j = \sum_{i=1}^k{P_{ij} a_i} \sim \mathcal{N}\left(0, \frac{\sum_{i=1}^k{a_i^2}}{q}\right) = \mathcal{N}\left(0, \frac{\|\mathbf{a}\|^2}{q} \right) \label{normal_distr}
\end{align}
Same applies for $(P\mathbf{b})_j$. We use this fact in the following derivation:
\begin{align}
    Var_P[P\mathbf{a} \cdot P\mathbf{b}] = \sum_{j=1}^q{Var_P\left[ \left(\sum_{i=1}^k{P_{ij} a_i}\right)\left(\sum_{i=1}^k{P_{ij} b_i}\right) \right]} \\
    = \sum_{j=1}^q \EX_P\left[ \left(\sum_{i=1}^k{P_{ij} a_i}\right)^2\left(\sum_{i=1}^k{P_{ij} b_i}\right)^2 \right] \nonumber - \sum_{j=1}^q \underbrace{\EX_P\left[ \left(\sum_{i=1}^k{P_{ij} a_i}\right)\left(\sum_{i=1}^k{P_{ij} b_i}\right) \right]^2}_{=(\mathbf{a} \cdot \mathbf{b})^2/q^2, \textrm{from (\ref{exp_q_begin}-\ref{exp_q_end})}} \\
    = \sum_{j=1}^q \EX_P\left[ \left(\sum_{i=1}^k{P_{ij} a_i}\right)^2\left(\sum_{i=1}^k{P_{ij} b_i}\right)^2 \right] - \sum_{j=1}^q\frac{(\mathbf{a} \cdot \mathbf{b})^2}{q^2} \label{inequality_1} \\
    \leq \sum_{j=1}^q{\sqrt{\EX_P\left[ \left( \sum_{i=1}^k{P_{ij} a_i} \right)^4 \right] \EX_P\left[ \left( \sum_{i=1}^k{P_{ij} b_i} \right)^4 \right]}} \nonumber - \frac{(\mathbf{a} \cdot \mathbf{b})^2}{q} \label{inequality_2} \\
    = \sum_{j=1}^q{\sqrt{\frac{3\|\mathbf{a}\|^4}{q^2} \cdot \frac{3\|\mathbf{b}\|^4}{q^2}}} - \frac{(\mathbf{a} \cdot \mathbf{b})^2}{q} \\
    = \sum_{j=1}^q{\frac{3\|\mathbf{a}\|^2 \|\mathbf{b}\|^2}{q^2}} - \frac{(\mathbf{a} \cdot \mathbf{b})^2}{q} \\
    = \frac{3\|\mathbf{a}\|^2 \|\mathbf{b}\|^2 - (\mathbf{a} \cdot \mathbf{b})^2}{q}
\end{align}
We used Cauchy–Schwarz inequality in (\ref{inequality_1}-\ref{inequality_2}). In (\ref{inequality_2}) under the square root 
there is a product of the 4th moments of two
normal distributions derived in (\ref{normal_distr}).

\section*{Appendix B: Comparison of generated data with ground truth}

\begin{figure}[!t]
\centering
\begin{minipage}{0.45\textwidth}
\centering
\includegraphics[width=0.9\textwidth]{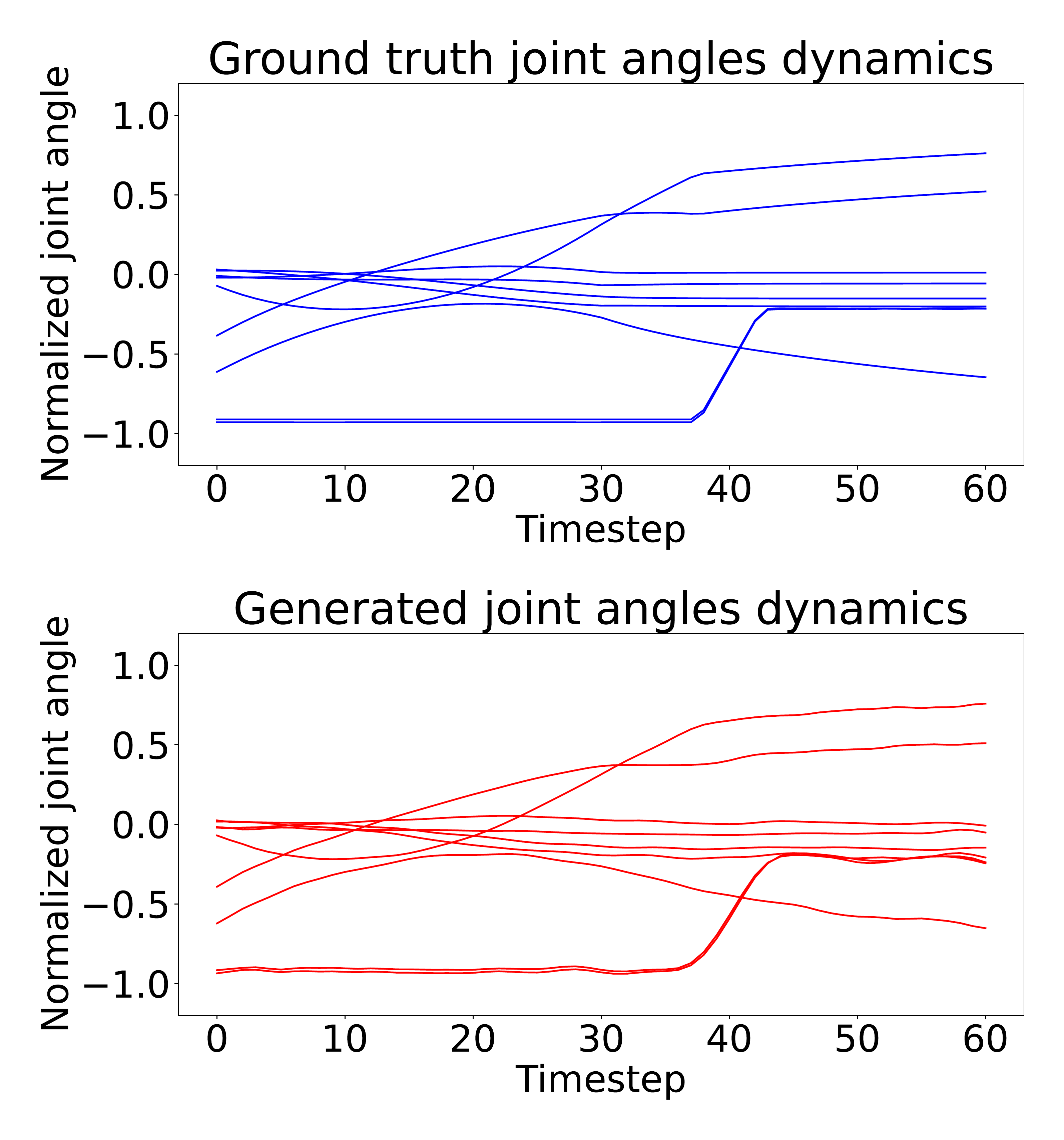}

"Grasp the block" primitive sample
\end{minipage}
\begin{minipage}{0.45\textwidth}
\centering
\includegraphics[width=0.9\textwidth]{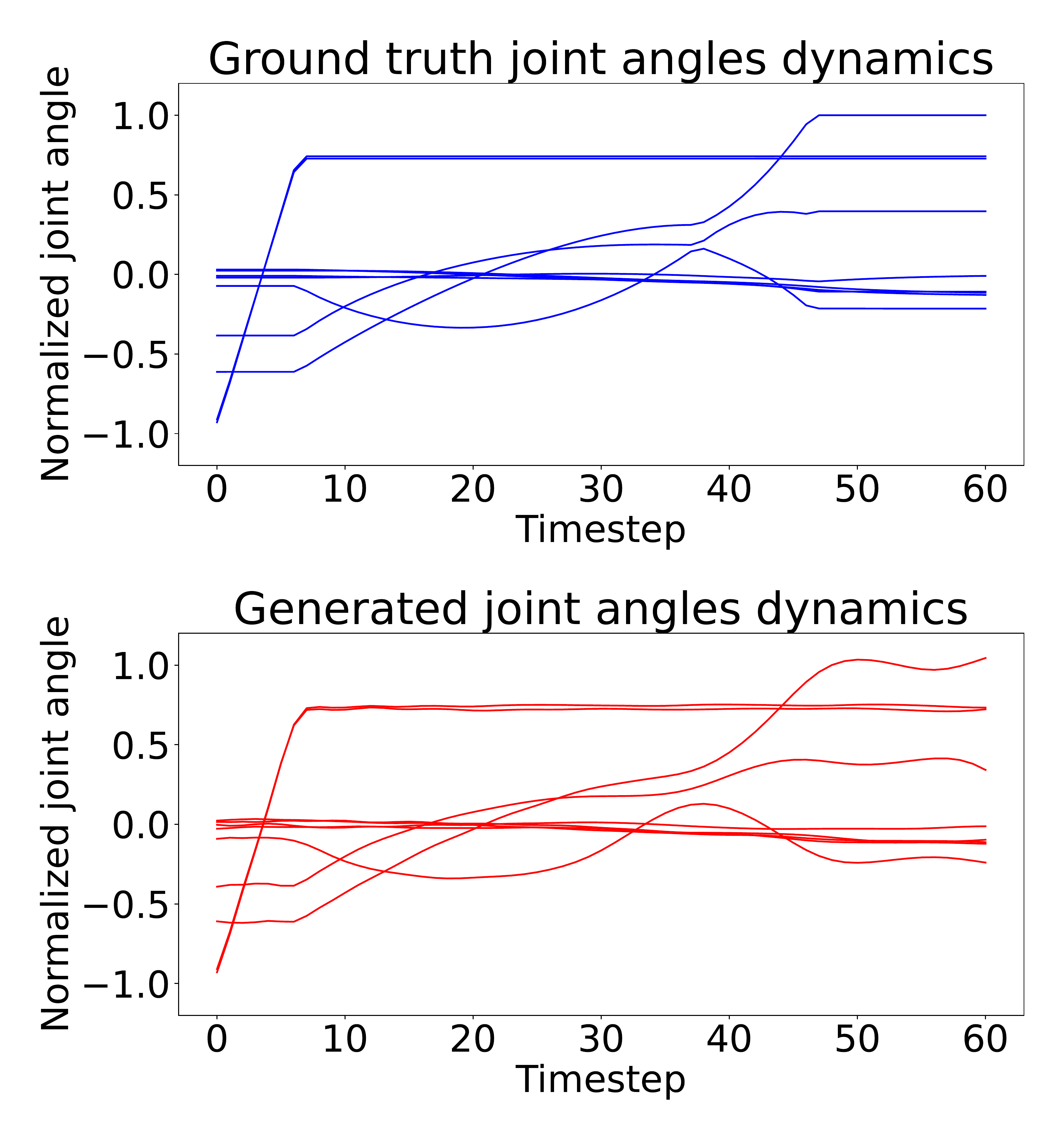}

"Push the block" primitive sample
\end{minipage}

\begin{minipage}{0.45\textwidth}
\centering
\includegraphics[width=0.9\textwidth]{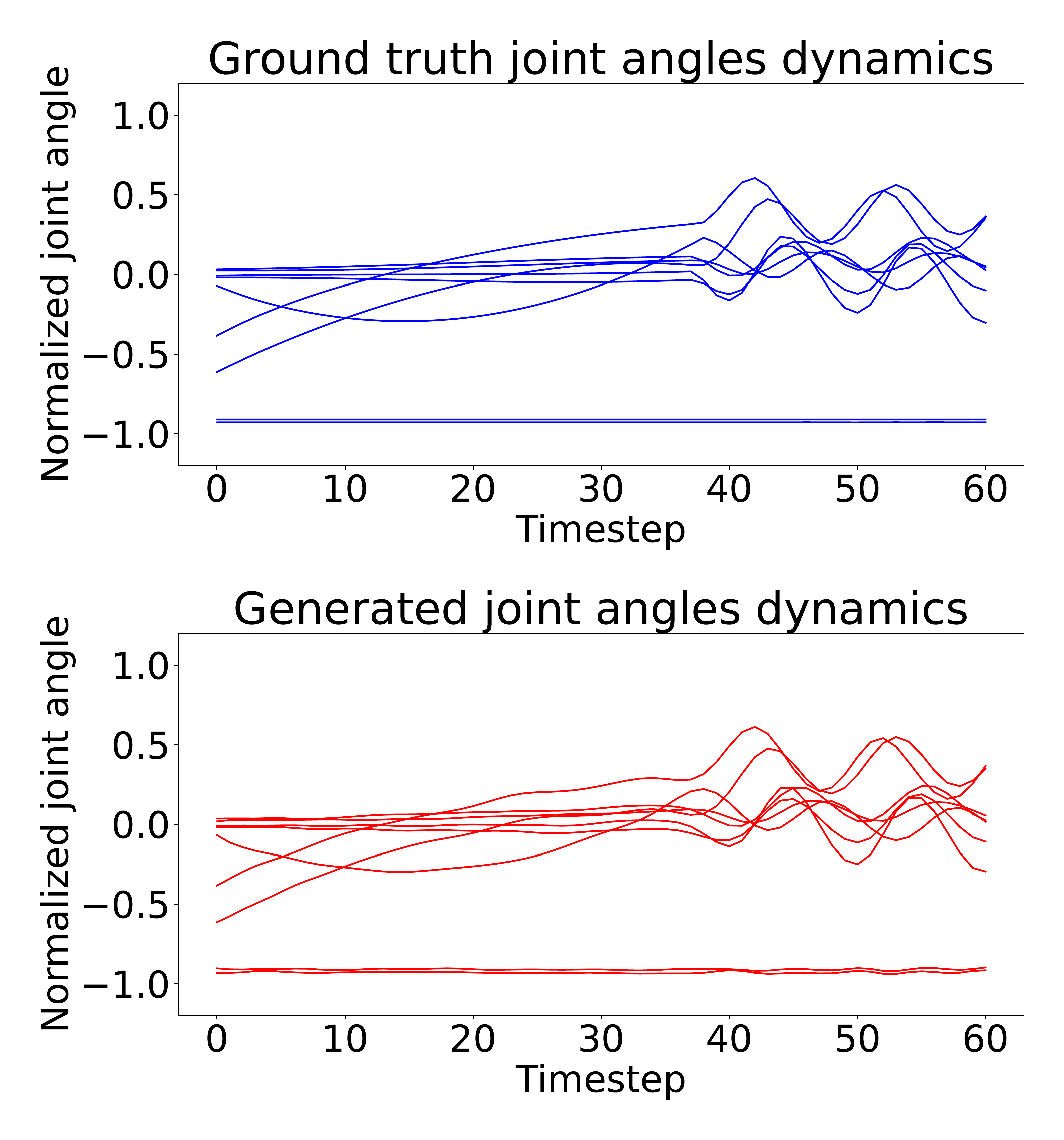}

"Circular motion around the block" primitive sample
\end{minipage}

\caption{Sequences of joint angles for seven joints and two fingers of the robotic arm 
generated by model and the ground truth. All models with sufficiently low loss function values
show similar results. This image is for the model with latent dimension equal 40 and random linear projection initialization method.}
\label{fig:joint_angles}
\end{figure}

Comparison of generated sequences of joint angles with ground truth is presented in Fig. \ref{fig:joint_angles}.
Examples of recorded image sequence together with prediction is available at the following link: https://doi.org/10.6084/m9.figshare.19235277.v1

\end{document}